\documentclass[10pt,twocolumn,letterpaper]{article}

\usepackage{cvpr}              

\usepackage{multirow}

\definecolor{cvprblue}{rgb}{0.21,0.49,0.74}
\usepackage[pagebackref,breaklinks,colorlinks,allcolors=cvprblue]{hyperref}
\usepackage{xspace}
\usepackage{array}
\newcommand{\name}{MoCapAnything\xspace}

\def\paperID{11070} 
\def\confName{CVPR}
\def\confYear{2026}

\title{MoCapAnything: Unified 3D Motion Capture for Arbitrary Skeletons from Monocular Videos}

\author{
Kehong Gong$^{*,1}$, Zhengyu Wen$^{*,2}$, Weixia He$^{2}$, Mingxi Xu$^{2}$, Qi Wang$^{2}$, Ning Zhang$^{2}$, \\ Zhengyu Li$^{2}$, Dongze Lian$^{2}$, Wei Zhao$^{2}$, Xiaoyu He$^{2}$, Mingyuan Zhang$^{\dagger,2}$ \\
$^{1}$Huawei International Pte. Ltd., $^{2}$Huawei Central Media Technology Institute \\
{\small $^{*}$ Equal Contributions, $^{\dagger}$ Corresponding Author} \\
}

\begin{document}

\newcolumntype{Y}{>{\centering\arraybackslash}X}
\twocolumn[{%
\renewcommand\twocolumn[1][]{#1}%
\maketitle
\begin{center}
  \vspace{-6 mm}
  \centering
  \captionsetup{type=figure}
  \vspace{-2 mm}
  \resizebox{\linewidth}{!}{
    \includegraphics{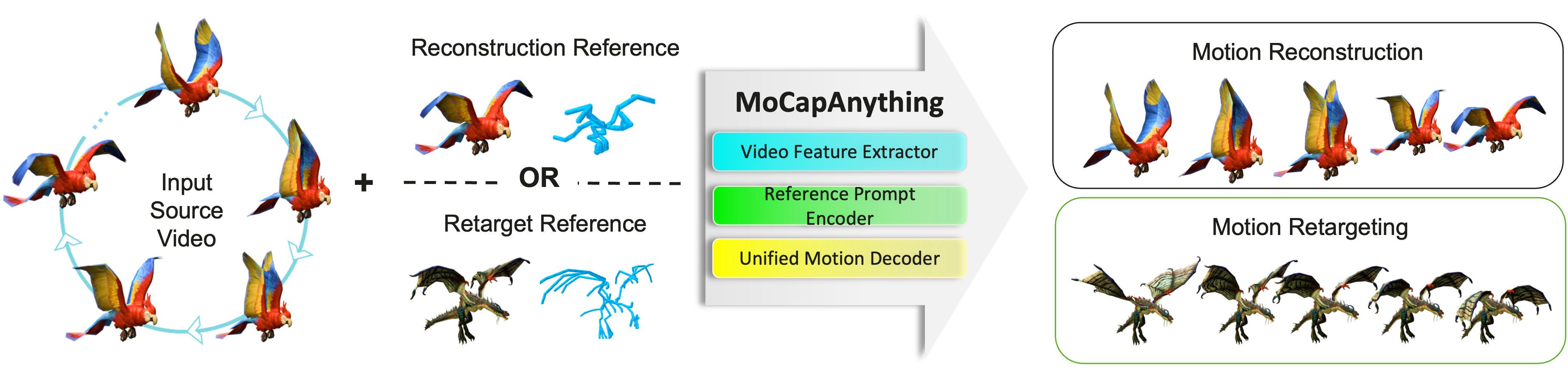}
  }
  \caption{Overview of our \name\ framework. Given a monocular video and a reference 3D asset (mesh/skeleton/appearance), our system first reconstructs a 4D mesh sequence from the video and encodes the reference asset via a multi-modal prompt encoder. A unified motion decoder then predicts joint trajectories, followed by an IK fitting stage that outputs animation in the asset’s own rig convention. The framework supports both direct motion capture (reference matches video subject) and cross-asset retargeting (reference differs from video).}
  \label{fig:teaser}
\end{center}%
}]

\begin{abstract}
Motion capture now underpins content creation far beyond digital humans, yet most pipelines remain species- or template-specific. We formalize this gap as Category-Agnostic Motion Capture (CAMoCap): given a monocular video and an arbitrary rigged 3D asset as a prompt, the goal is to reconstruct a rotation-based animation (e.g., BVH) that directly drives the specific asset. We present \name, a reference-guided, factorized framework that first predicts 3D joint trajectories and then recovers asset-specific rotations via constraint-aware Inverse Kinematics (IK) Fitting. \name comprises three learnable modules and a lightweight IK stage: a \textbf{Reference Prompt Encoder} that distills per-joint queries from the asset’s skeleton, mesh, and rendered image set; a \textbf{Video Feature Extractor} that computes dense visual descriptors and reconstructs a coarse 4D deforming mesh to bridge the modality gap between RGB tokens and the point-cloud–like joint space; and a \textbf{Unified Motion Decoder} that fuses these cues to produce temporally coherent trajectories. We also curate Truebones Zoo~\cite{truebones_mocap} with 1{,}038 motion clips, each providing a standardized skeleton–mesh–rendered-video triad. Experiments on in-domain benchmarks and in-the-wild videos show that \name delivers high-quality skeletal animations and exhibits non-trivial cross-species retargeting across heterogeneous rigs, offering a scalable path toward prompt-based 3D motion capture for arbitrary assets. The code is available on our project page: \url{https://animotionlab.github.io/MoCapAnything/}

\end{abstract}    
\section{Introduction}
\label{sec:intro}

Motion capture underpins modern content creation beyond digital humans, yet most pipelines remain tied to a single species or template. Human-centric systems typically regress SMPL-family~\citep{loper2015smpl,pavlakos2019expressive} parameters from monocular inputs (e.g., DeepPose~\citep{Toshev_2014_CVPR} for 2D keypoints and HMR~\citep{kanazawa2018end} for SMPL-based 3D recovery) and work well only within that fixed topology. For non-human subjects, category-agnostic keypoint detection (CAPE) broadens 2D landmark coverage via promptable support examples, but it stops short of producing animation-ready 3D motion~\citep{rusanovsky2025capex}. On the motion side, animal mocap usually builds on SMAL~\citep{Zuffi:CVPR:2017} and is limited to a few quadruped categories, with models and rig assumptions that do not transfer to diverse assets. Consequently, existing solutions fall short in practical pipelines where creators must (i) retarget human/animal motion to non-biological rigs (robots, mechs, toys, articulated props), (ii) animate large heterogeneous asset libraries for games and crowd scenes, (iii) drive VTuber/virtual-production avatars that frequently change topology, and (iv) spin up IP-specific characters (mascots, creatures) without building a new parametric model per species. 

To address the limitations of fixed-species motion capture, we recast the problem as prompt-based 3D motion capture: given a monocular video and an arbitrary rigged 3D asset, the goal is to reconstruct a rotation-based animation (e.g., BVH joint rotations) that directly drives that specific character. We refer to this setting as Category-Agnostic Motion Capture (CAMoCap). To make this concrete and reproducible, we curate the Truebones Zoo benchmark, where each motion instance provides a clean bundle comprising the rigged skeleton (with standardized joint names and hierarchy), the mesh, and an asset-aligned rendered video. The dataset contains 1,038 motion clips. We hold out 60 for the test set and use the remaining 978 for training.

CAMoCap raises three core challenges. First, \textit{motion representation}: joint rotations are defined in asset-local frames, so direct angle regression across diverse rest poses is brittle. Second, \textit{reference-guided estimation}: the model must inject information about the target asset into video-based 3D keypoint prediction effectively. Third, \textit{multimodal integration}: there is a gap between dense RGB features and the point-cloud–like structure of keypoints. Bridging them naively may lead to suboptimal accuracy.

To tackles above mentioned challenges, we propose a novel framework, \name, which factorizes motion recovery into (i) 3D keypoint trajectory prediction and (ii) per-joint rotation recovery. It uses three learnable modules followed by a lightweight IK stage. The \textbf{Reference Prompt Encoder} distills the asset’s mesh, skeleton, and rendered image set into structure-aware per-joint queries. The \textbf{Video Feature Extractor} computes dense visual descriptors (e.g., DINOv2~\cite{DBLP:journals/corr/abs-2304-07193}) and reconstructs a coarse 4D deforming mesh from the input video. This mesh contributes topology- and geometry-aware cues that bridge the gap between RGB tokens and the point-cloud nature of joints. The \textbf{Unified Motion Decoder} attends over reference queries and video features to produce temporally consistent 3D joint trajectories. Finally, IK Fitting converts these trajectories into asset-specific rotations while respecting hierarchy, bone lengths, joint limits, and temporal smoothness. This modular factorization naturally supports both motion capture (same skeleton) and retargeting (different skeletons) across heterogeneous rigs.

Our main contributions are summarized as follows:
\begin{enumerate}
\item We formalize a new task, \emph{Category-Agnostic Motion Capture} (CAMoCap), prompt-based 3D motion capture from a monocular video and an arbitrary rigged 3D asset. We also release reorganized Truebones Zoo~\cite{truebones_mocap} with 1{,}038 clips, each providing a  skeleton–mesh–rendered-video triad.
\item We present the first framework for CAMoCap, \name, to yield temporally coherent, animation-ready results across heterogeneous rigs. Specifically, we decouple motion into 3D joint trajectories followed by IK-based rotations to stablize training process and introduce mesh as an auxiliary modality to bridge RGB tokens and joint space.
\item \name attains strong in-domain accuracy, generalizes to in-the-wild videos, and shows non-trivial cross-species mocap and retargeting.
\end{enumerate}

\section{Related Works}
\label{sec:related}
\subsection{Pose Estimation}
Human 2D pose estimation aims to localize anatomical keypoints in images. Classic methods are typically grouped into bottom-up and top-down paradigms: bottom-up approaches first detect all keypoints and then group them into person instances~\citep{cheng2020bottom}, while top-down pipelines detect person bounding boxes and run a single-person pose head on each crop~\citep{xiao2018simple}. Within the top-down family, heatmap-based networks such as Stacked Hourglass~\citep{newell2016stacked}, CPN~\citep{chen2018cascaded}, SimpleBaseline~\citep{xiao2018simple}, HRNet~\citep{sun2019deep}, Simple Pose~\citep{li2020simple}, and ViTPose~\citep{xu2022vitpose} predict per-joint likelihood maps from multi-scale or high-resolution features, whereas regression-style methods including DeepPose~\citep{toshev2014deeppose}, RLE~\citep{li2021human}, and SimCC~\citep{li2022simcc} directly output coordinates or 1D classifications to alleviate heatmap quantization. More recent DETR-style frameworks~\citep{carion2020end} treat poses and/or keypoints as query sets and perform end-to-end multi-person estimation without hand-crafted grouping~\citep{shi2022end,xiao2022querypose,yang2023explicit}, and vision–language approaches such as LocLLM~\citep{wang2024locllm} encode keypoints as text descriptions to enable some zero-shot generalization to new landmarks; however, all these architectures remain tightly coupled to a predefined human skeleton~\citep{gong2021poseaug,gong2022posetriplet} and keypoint set.

Beyond these category-specific keypoint detectors, an emerging line of work aims to relax the dependence on fixed object categories through category-agnostic pose estimation (CAPE). CAPE formulates pose estimation as a few-shot problem, where a single model predicts keypoints for unseen categories by comparing support keypoints with query images in a shared embedding space~\citep{xu2022pose}. POMNet~\citep{xu2022pose} instantiates CAPE with a transformer encoder over query images and support keypoints, and regresses similarity scores from their concatenated features. CapeFormer~\citep{shi2023matching} further adopts a two-stage matching framework that first proposes candidate correspondences and then refines unreliable matches to improve localization accuracy. Pose Anything~\citep{hirschorn2023graph} departs from treating keypoints as isolated entities and instead models them as nodes in a graph, using graph convolutions to exploit structural relationships, break symmetries, and better handle occlusions. More recently, CapeX~\citep{rusanovsky2024capex} pushes CAPE beyond purely visual correspondence by replacing annotated support images with text prompts attached to graph nodes, aligning query image features to open-vocabulary textual keypoint descriptions. While these CAPE methods significantly improve generalization across categories, they operate in 2D and focus on static keypoint localization, without modeling 3D trajectories, temporal consistency, or animation-ready joint representations, which are central to our monocular motion capture setting.

\subsection{Motion Capture}

Monocular human motion capture is typically formulated as recovering pose and shape parameters of parametric whole-body models such as SMPL~\cite{loper2015smpl} and SMPL-X~\citep{pavlakos2019expressive}. With whole-body models, expressive human pose and shape (EHPS) estimation from a single RGB image or video—jointly modeling body, hands, and face—has attracted much attention. Early optimization-based methods (e.g., SMPLify-X~\citep{pavlakos2019expressive}) fit SMPL-X to detected 2D keypoints but are slow and brittle. 

One-stage frameworks such as OSX~\citep{lin2023one}, AiOS~\citep{sun2024aios}, and MultiHMR~\citep{baradel2024multi} instead use ViT-style backbones to jointly localize and regress full SMPL-X parameters in a single forward pass, alleviating error accumulation and improving robustness. Beyond image-aligned meshes, recent work distinguishes between camera-space and world-grounded human motion recovery. Most HMR and video-based approaches follow the camera-space formulation, regressing SMPL parameters directly from images or clips with CNN, RNN, or transformer encoders (e.g., HMR~\citep{kanazawa2018end}/HMR2.0~\citep{goel2023humans}, VIBE~\citep{kocabas2020vibe}, TCMR~\citep{choi2021beyond}), which yields accurate pose but entangles motion with camera movements. To obtain physically meaningful trajectories, multi-camera studios and IMU-based systems rely on calibration or inertial sensors, while recent monocular methods integrate SLAM or visual odometry with learned motion priors (e.g., SLAHMR~\citep{yuan2022glamr}, PACE~\citep{kocabas2024pace}, TRAM~\citep{wang2024tram}, WHAC~\citep{yin2024whac}, WHAM~\citep{shin2024wham}) to estimate global motion. However, these pipelines remain tied to a single human template and are difficult to extend to more general, non-human skeletons.

Beyond humans, 3D animal reconstruction has been explored under two main paradigms: model-free and model-based. Model-free methods make minimal assumptions about anatomy and directly recover a deformable surface, e.g., CMR~\citep{kanazawa2018learning} deforms a spherical template to reconstruct birds, while LASSIE~\citep{yao2022lassie}, MagicPony~\citep{wu2023magicpony}, and 3D-Fauna~\citep{li2024learning} learn articulated 3D shape from image collections; ViSER~\citep{yang2021viser}, LASR~\citep{yang2021lasr}, BANMo~\citep{yang2022banmo}, and PPR~\citep{yang2023ppr} extend this idea to videos. In contrast, model-based approaches assume a species-specific or parametric 3D template is given or retrievable~\citep{wu2022casa}, enabling pose- and shape-aware analysis over time. SMAL~\citep{Zuffi:CVPR:2017}, an articulated quadruped model learned from toy scans, has been widely used~\citep{DBLP:journals/corr/abs-2412-08101}. However, these pipelines remain species- and template-specific, and do not generalize to the diverse, non-animal skeletons required by arbitrary animatable assets.
\section{Method}
\label{sec:method}

\begin{figure*}[t]
\centering
\includegraphics[width=\linewidth]{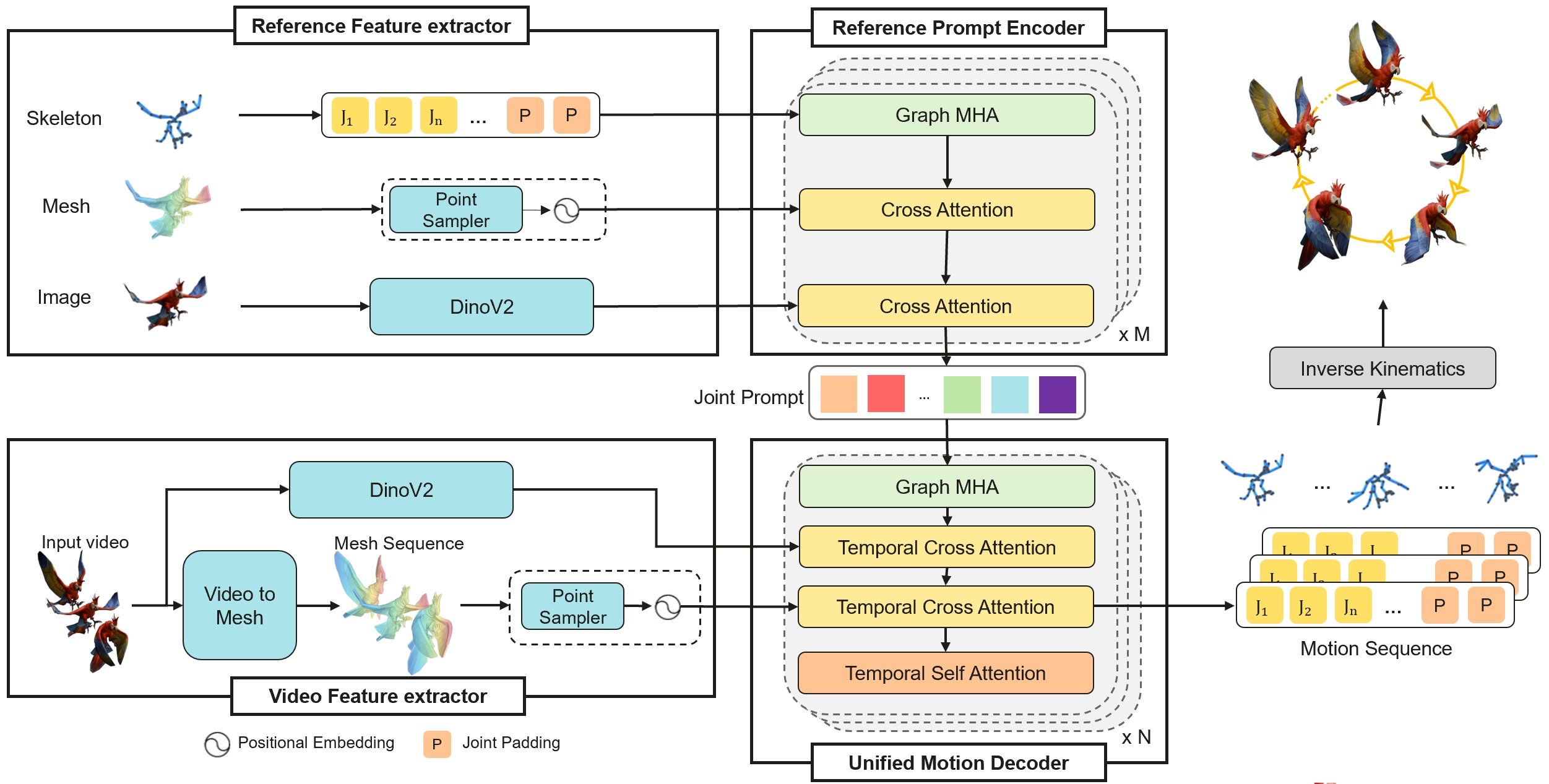}
\caption{
Detailed architecture of our method. A multi-modal Reference Prompt Encoder fuses mesh, skeleton, and appearance of the target asset into per-joint queries. A monocular video is converted into a 4D mesh sequence, and both mesh and video features are extracted. The Unified Motion Decoder fuses these signals via multi-branch attention to predict 3D keypoints, which are converted to asset-specific joint rotations via an optimization-based IK layer.
}
\vspace{-5pt}
\label{fig:framework}
\end{figure*}

\subsection{Task Formulation}

In this work, we propose a new task, \textbf{Category-Agnostic Motion Capture} (CAMoCap), which aims to reconstruct motion for arbitrary 3D assets with diverse skeletal topologies from monocular videos. This formulation transcends traditional paradigms centered on human or category-specific mocap, enabling both motion capture and retargeting for assets of any type or skeletal structure, and thus brings broader applicability and flexibility to animation, virtual production, and creative content creation.

Formally, given a monocular RGB video $V=\{I_t\}_{t=1}^T$ depicting a moving character or creature, and a rigged 3D asset $A=(\mathcal{M},\mathcal{S},\mathcal{I}_A)$ with arbitrary skeletal structure, the goal is to predict a sequence of joint rotations $\{\mathbf{R}_t\}_{t=1}^T$ that animates $A$ in accordance with the motion in $V$:
\begin{equation}
(V,A)\ \longrightarrow\ \{\mathbf{R}_t\}_{t=1}^T,\ 
\mathbf{R}_t=\{R_{t,j}\}_{j\in\mathcal{J}},\ \ R_{t,j}\in\mathrm{SO}(3).
\end{equation}
Here, $\mathcal{M}$ denotes the mesh, and the skeleton is $\mathcal{S}=(\mathcal{J},\mathcal{E},\mathbf{o})$, where $\mathcal{J}$ is the joint set and $\mathcal{E}\subseteq\mathcal{J}\times\mathcal{J}$ denotes the directed parent$\rightarrow$child edges. For each edge $(i\!\rightarrow\!j)\in\mathcal{E}$, $\mathbf{o}_{i\rightarrow j}\in\mathbb{R}^3$ is the offset of joint $j$ relative to its parent $i$. The rest rig also specifies canonical joint labels via a naming function $\ell:\mathcal{J}\to\mathcal{N}$. The optional appearance is provided as a reference image set $\mathcal{I}_A=\{I_A^{(k)}\}_{k=1}^{K}$ (e.g., renders or photos of $A$). This general formulation covers both \emph{motion capture} (when $A$ matches the subject in $V$) and \emph{motion retargeting} (when $A$ differs from the subject).

\subsection{Overview}

To tackle the CAMoCap task, we employ three dedicated, learnable branches to extract features from the reference prompt and the input monocular video, fuse them, and estimate motion sequences. A naive alternative is to regress joint rotations directly after feature fusion, but in monocular settings this is brittle due to: (i) parameterization and rig–frame ambiguities that make angles asset-dependent, (ii) under-constrained evidence where depth and camera motion entangle local rotations, and (iii) poor temporal continuity from per-frame angle regression. We therefore decompose the problem into 3D keypoint trajectory estimation followed by rotation recovery via inverse kinematics (IK).

Accordingly, our approach, \name, comprises four components (see Fig.~\ref{fig:framework}):
\begin{enumerate}
    \item \textbf{Reference Prompt Encoder}: Extracts per-joint features from the reference asset, including skeletal, mesh, and appearance image-set cues.
    \item \textbf{Video Feature Extractor}: Uses off-the-shelf models to obtain visual descriptors (e.g., DINOv2~\cite{DBLP:journals/corr/abs-2304-07193}) and reconstruct a coarse 4D deforming mesh from the video. The mesh supplies topology- and geometry-aware signals that bridge the modality gap between dense visual tokens and the point-cloud–like joint representation, stabilizing and improving trajectory estimation.
    \item \textbf{Unified Motion Decoder}: Fuses reference, geometric, and visual information to predict temporally coherent 3D joint trajectories for the target asset.
    \item \textbf{IK Fitting Process}: Converts predicted joint trajectories into asset-specific joint rotations via an optimization-based IK procedure that respects hierarchy, bone lengths, joint limits, and temporal smoothness.
\end{enumerate}
This modular pipeline flexibly supports both motion capture (same skeleton) and retargeting (different skeletons) for arbitrary 3D assets and rig topologies.

\subsection{Architecture Design}
\label{sec:arch}

\subsubsection{Reference Prompt Encoder}

Let the reference asset be $A=(\mathcal{M},\mathcal{S},\mathcal{I}_A)$ with mesh $\mathcal{M}$, skeleton $\mathcal{S}=(\mathcal{J},\mathcal{E},\mathbf{o})$, and an image set $\mathcal{I}_A=\{I_A^{(k)}\}_{k=1}^{K}$. The encoder outputs per-joint queries $Q=\{\,\mathbf{q}_j\in\mathbb{R}^d\,\}_{j\in\mathcal{J}}$. For each joint $j$ with coordinate $\mathbf{x}_j\in\mathbb{R}^3$, we apply a sinusoidal positional encoding $\mathrm{pe}(\mathbf{x}_j)$ and a linear projection to obtain an initial embedding
$
\mathbf{q}_j^{(0)}=\mathbf{W}_p[\mathrm{pe}(\mathbf{x}_j);\ \mathbf{e}_{\text{name}}(\ell(j))]+\mathbf{b}_p
$
(where $\mathbf{e}_{\text{name}}$ is optional). Variable joint counts are handled by a binary mask $\mathbf{m}\in\{0,1\}^{|\mathcal{J}_{\max}|}$ that zeroes padded joints in all attention operations.

We then apply $L$ stacked fusion blocks, each with three submodules in a row:

\begin{enumerate}
  \item \textbf{Self-Attention with Skeleton Topology.}
  We use a graph multi-head attention (Graph-MHA) on $\{\mathbf{q}_j^{(\ell)}\}$ with an attention bias $\mathbf{B}_{ij}$ computed from skeleton topology (adjacency in $\mathcal{E}$ and geodesic/kinematic distances), following the AnyTop~\cite{gat2025anytop} design:
  \begin{equation}
  \begin{split}
  \mathrm{Attn}(\mathbf{q}_i,\mathbf{q}_j)\propto\frac{\langle \mathbf{W}_Q\mathbf{q}_i,\ \mathbf{W}_K\mathbf{q}_j\rangle}{\sqrt{d}}+\mathbf{B}_{ij},\\ \mathbf{B}_{ij}=f_{\text{topo}}(\mathcal{E}, i, j).
  \end{split}
  \end{equation}
  This encourages structure-aware message passing along the kinematic tree. Details will be illustrated in the supplementary material.

  \item \textbf{Cross-Attention to Mesh Geometry.}
  We sample surface points from $\mathcal{M}$ to form $\mathcal{P}=\{(\mathbf{p}_u,\mathbf{n}_u)\}_{u=1}^{U}$ (positions and normals). Mesh tokens are
  $
  \mathbf{g}_u=\mathbf{W}_m[\mathrm{pe}(\mathbf{p}_u);\ \mathbf{n}_u].
  $
  Joints attend to $\{\mathbf{g}_u\}$ to learn implicit skinning-like relations between joints and local surface geometry.

  \item \textbf{Cross-Attention to Appearance.}
  Images in $\mathcal{I}_A$ are encoded by a frozen image encoder $\phi_{\text{img}}$ (e.g., DINOv2) to obtain appearance tokens $\phi_{\text{img}}(\mathcal{I}_A)$. We inject appearance cues that disambiguate symmetric or visually similar parts via cross-attention mechanism.
\end{enumerate}

Across layers, masked attention ensures invariance to the absolute joint count, and residual/FFN updates refine $\mathbf{q}_j^{(\ell)}\!\to\!\mathbf{q}_j^{(\ell+1)}$ by progressively integrating \emph{structural} ($\mathcal{S}$), \emph{geometric} ($\mathcal{M}$), and \emph{visual} ($\mathcal{I}_A$) evidence. The final per-joint queries $Q=\{\mathbf{q}_j^{(L)}\}$ serve as asset-specific prompts for the Unified Motion Decoder and enable robust generalization across diverse characters and skeleton topologies.

\subsubsection{Video Feature Extractor}
Given a monocular video $V=\{I_t\}_{t=1}^{T}$, we build two complementary streams.

\textbf{Visual stream.} Each frame is encoded by a frozen DINOv2 image encoder $\phi_{\text{img}}$, yielding per-frame dense tokens $\mathbf{A}_t=\phi_{\text{img}}(I_t)$ (and an optional global token). These serve as appearance/texture cues.

\textbf{Geometry stream.} We apply a pretrained image-to-3D reconstructor to obtain a coarse deforming surface sequence $\widehat{\mathcal{M}}=\{\widehat{\mathcal{M}}_t\}_{t=1}^{T}$. For each $t$, we randomly downsample the surface to $U{=}1024$ points $\mathcal{P}_t=\{(\mathbf{p}_{t,u},\mathbf{n}_{t,u})\}_{u=1}^{U}$. Points are embedded as
\[
\mathbf{g}_{t,u}=\mathbf{W}_m\,[\mathrm{pe}(\mathbf{p}_{t,u});\,\mathbf{n}_{t,u};\,\mathrm{pe}(t)],
\]
producing geometry-aware tokens $\mathbf{G}_t=\{\mathbf{g}_{t,u}\}_{u=1}^{U}$ analogous to the mesh features used in the Reference Prompt Encoder.

We form the video feature set $\mathcal{V}=\{\mathbf{A}_t,\mathbf{G}_t\}_{t=1}^{T}$ (keys/values for the decoder). The 4D mesh tokens provide topology/geometry signals that bridge dense RGB features and the point-cloud–like joint space, stabilizing subsequent 3D keypoint estimation.

\subsubsection{Unified Motion Decoder}
Given the per-joint prompts $Q=\{\mathbf{q}_j\}_{j\in\mathcal{J}}$, the skeleton $\mathcal{S}=(\mathcal{J},\mathcal{E})$, and video features $\mathcal{V}=\{\mathbf{A}_t,\mathbf{G}_t\}_{t=1}^{T}$ (DINOv2-based visual tokens $\mathbf{A}_t$ and 4D-mesh point tokens $\mathbf{G}_t$), we tile $Q$ across time, add a temporal encoding to obtain $\{\mathbf{h}_{t,j}^{(0)}\}$, and apply a binary joint mask to accommodate variable-size skeletons. Each decoder layer refines these tokens through the following four stages:

\begin{enumerate}
    \item \textbf{Graph-based self-attention (intra-frame).} Within each frame, joint tokens are updated using an attention layer with an explicit topology bias derived from $\mathcal{E}$ (same as AnyTop~\citep{gat2025anytop}), ensuring that updates respect the kinematic tree and local limb couplings.
    \item \textbf{Temporal video cross-attention.} For each joint at time $t$, a sliding window over neighboring frames provides visual tokens that supply short-range appearance cues. Attending to this window improves continuity, fills in details under occlusion or motion blur, and stabilizes rapid movements.
    \item \textbf{Temporal point-cloud cross-attention.} Joint tokens then aggregate geometry-aware evidence from the corresponding 4D mesh window. These point tokens inject topology/shape signals that bridge dense RGB features and the point-cloud–like joint space, disambiguating depth and self-occlusion and capturing non-rigid deformations.
    \item \textbf{Temporal self-attention (per joint).} Finally, a windowed self-attention along the time axis mixes each joint’s past and future states to enforce longer-range consistency and reduce jitter, while better modeling higher-order dynamics.
\end{enumerate}

Residual connections, normalization, and feed-forward updates follow each block, and stacking $L$ layers progressively integrates \emph{structural} ($\mathcal{S}$), \emph{visual} ($\mathbf{A}_t$), and \emph{geometric} ($\mathbf{G}_t$) cues. A lightweight MLP head then predicts per-frame joint positions $\{\widehat{\mathbf{x}}_{t,j}\in\mathbb{R}^3\}$, yielding trajectories for the subsequent IK stage.

\subsection{Training Objective}
\label{sec:loss}

We supervise the decoder with a masked position regression loss consistent with our notation above. Let $\widehat{\mathbf{x}}_{t,j}\in\mathbb{R}^3$ be the predicted 3D position of joint $j\in\mathcal{J}$ at time $t\in\{1,\dots,T\}$, and let $\mathbf{x}_{t,j}$ be the ground-truth position. Since assets have different skeleton sizes, we pad all sequences to
$|\mathcal{J}_{\max}|$ joints and use a binary joint-validity mask
$\mathbf{m}\in\{0,1\}^{|\mathcal{J}_{\max}|}$ (with $m_j=1$ iff $j\in\mathcal{J}$ for this asset).

\[
\mathcal{L}_{\text{pos}}
=
\frac{1}{\sum_{t=1}^{T}\sum_{j} m_j}
\sum_{t=1}^{T}\sum_{j} m_j\,
\big\|\widehat{\mathbf{x}}_{t,j}-\mathbf{x}_{t,j}\big\|_1.
\]

We do not apply rotation space or explicit temporal losses during training: the network predicts joint positions, and rotations are obtained afterwards by the IK stage.

\subsection{IK Fitting Process}
\label{sec:ik}
We recover joint rotations from the predicted 3D joint trajectories using a lightweight two-stage IK procedure. First, we compute a per-frame geometric IK initialization by aligning rest-pose bone directions with the observed joint positions along each kinematic chain. This closed-form step provides a stable rotation estimate that respects the skeleton hierarchy. Then, we refine the rotations with a small differentiable IK optimization that minimizes the discrepancy between FK-reconstructed joints and the predicted 3D positions, while regularizing the solution toward the geometric initialization. The optimization is warm-started from the previous frame to ensure temporal stability and suppress unnecessary twist. This hybrid strategy produces accurate and smooth joint rotations at minimal computational cost. Additional implementation details are provided in the supplementary material.

\section{Experiments}
\label{sec:experiments}

\subsection{Dataset and Evaluation Protocol}

\begin{figure*}[t]
    \centering
    \includegraphics[width=\textwidth]{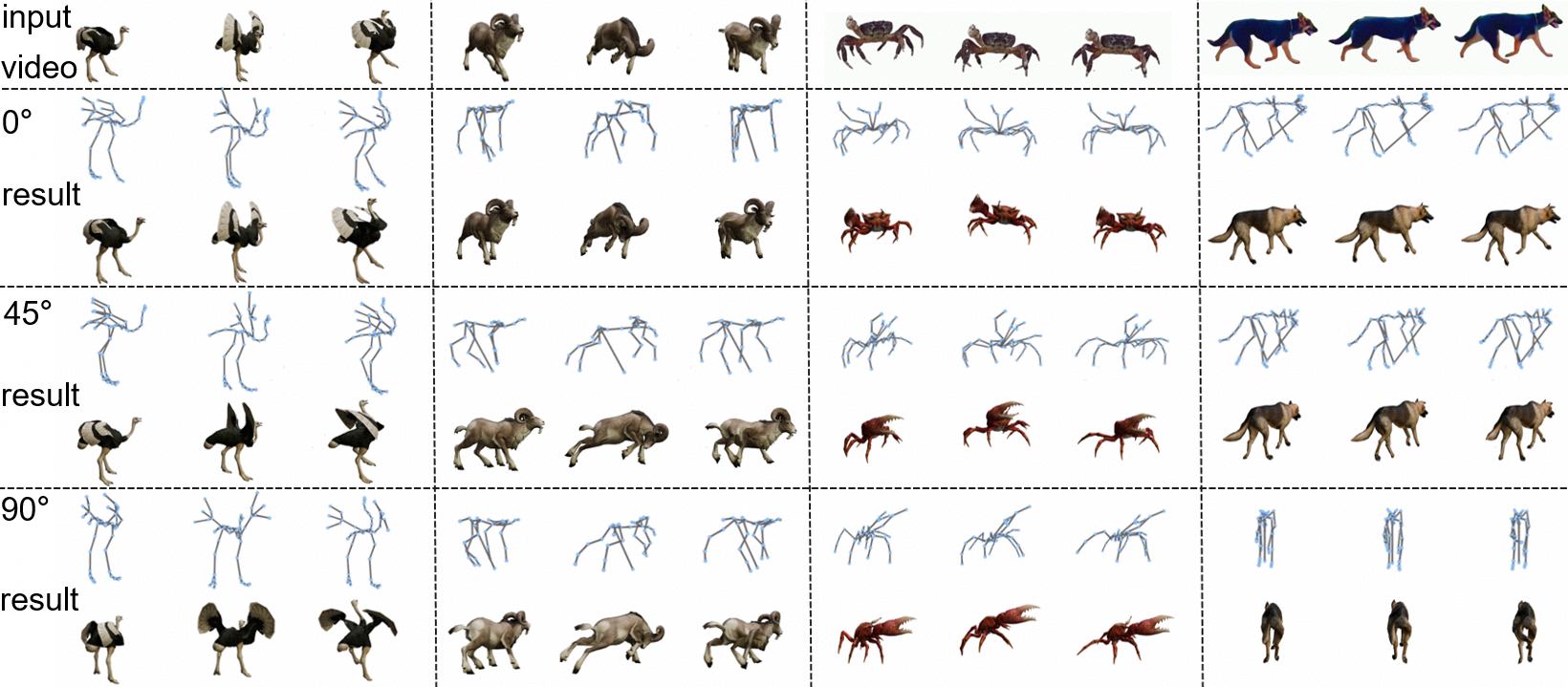}
    \caption{\textbf{IK-Driven Asset Animation} by ostrich (biped), a goat (quadruped), and in-the-wild cases such as a crab and a dog. We visualize the predicted poses and corresponding IK-driven animations from multiple viewpoints.}
    \label{fig:ik-result}
\end{figure*}

We evaluate our approach on the \textbf{Truebones Zoo}~\cite{truebones_mocap} dataset, which contains 1,038 animal motion sequences (totaling 104,715 frames) spanning a broad range of species and kinematic structures and 1000 random samples from objaverse~\cite{objaverse, objaverseXL}. For testing, we curate a set of 60 sequences with enough diversity, stratified into three groups: \textbf{Seen} (species with abundant training data), \textbf{Rare} (species with limited training data), and \textbf{Unseen} (species never seen during training). This protocol enables a thorough assessment of model generalization.

\subsection{Evaluation Metrics}

To disentangle the contributions of the two stages, we evaluate them separately: (i) 3D joint positions $\widehat{\mathbf{x}_{t,j}}$ and (ii) joint rotations $\widehat{R_{t,j}}$. In the main paper we focus on quantitative results for 3D keypoint prediction, while rotation-level evaluation (after IK) is deferred to the supplementary material. As for 3D keypoints, we report the following metrics:
\begin{itemize}
    \item \textbf{MPJPE} (Mean Per Joint Position Error): the mean Euclidean distance between predicted and ground-truth joint positions (lower is better).
    \item \textbf{MPJVE} (Mean Per Joint Velocity Error): the average velocity difference per joint, capturing temporal consistency and motion plausibility.
\end{itemize}

To account for large inter-species scale variations, we normalize all samples to a $[-1,1]^3$ cube for training and rescale both predictions and ground truth to a unified $1\mathrm{m}^3$ cube for evaluation. All metrics are reported in centimeters (cm). For quantitative evaluation, we use ground-truth meshes to compute joint positions for both training and testing data to avoid interference from predicted mesh quality, while all visualizations are based on predicted meshes.


\subsection{Compare with Baseline}

We compare with adapted baselines from established pose and motion capture pipelines, including ViTPose~\cite{xu2022vitpose}, HRNet~\cite{sun2019deep}, VIBE~\cite{kocabas2020vibe}, and GLoT~\cite{shen2023global}.
All baselines are re-trained on the same training set with unified skeleton representations and input modalities. 
For methods originally designed for human pose estimation, we adapt their output layers to match the target skeleton topology.
As shown in Table~\ref{tab:baseline-mpjpe-mpjve}, our method consistently outperforms all baselines across seen, rare, and unseen splits, demonstrating the effectiveness of our category-agnostic design.

\begin{table}[t]
\centering
\normalsize
\setlength{\tabcolsep}{3pt}
\resizebox{\columnwidth}{!}{
\begin{tabular}{lcccccc}
\toprule
& \multicolumn{2}{c}{\textbf{Seen}} & \multicolumn{2}{c}{\textbf{Rare}} & \multicolumn{2}{c}{\textbf{Unseen}} \\
\cmidrule(lr){2-3} \cmidrule(lr){4-5} \cmidrule(lr){6-7}
\textbf{Method} & 
\textbf{MPJPE $\downarrow$} & \textbf{MPJVE $\downarrow$} & 
\textbf{MPJPE $\downarrow$} & \textbf{MPJVE $\downarrow$} & 
\textbf{MPJPE $\downarrow$} & \textbf{MPJVE $\downarrow$} \\
\midrule
HRNet   & 9.77 & 1.41 & 10.86 & 1.55 & 23.53 & 1.84 \\
ViTPose & 9.19 & 1.73 & 9.33  & 1.40 & 23.37 & 2.15 \\
VIBE    & 4.46 & 0.83 & 4.03  & 0.68 & 8.72  & 0.95 \\
GLoT    & 3.98 & 1.37 & 3.58  & 0.84 & 7.42  & 2.18 \\
\textbf{Ours} 
        & \textbf{1.06} & \textbf{0.44} 
        & \textbf{1.28} & \textbf{0.37} 
        & \textbf{1.76} & \textbf{0.36} \\
\bottomrule
\end{tabular}
}
\caption{Comparison with baseline methods on the Truebones Zoo-test set. Lower is better ($\downarrow$). Results are reported across three generalization levels: seen, rare, and unseen species.}
\label{tab:baseline-mpjpe-mpjve}
\end{table}

\begin{table}[t]
\centering
\normalsize
\setlength{\tabcolsep}{3pt}
\resizebox{\columnwidth}{!}{
\begin{tabular}{lcccccc}
\toprule
& \multicolumn{2}{c}{\textbf{Seen}} & \multicolumn{2}{c}{\textbf{Rare}} & \multicolumn{2}{c}{\textbf{Unseen}} \\
\cmidrule(lr){2-3} \cmidrule(lr){4-5} \cmidrule(lr){6-7}
\textbf{Method} & 
\textbf{MPJPE $\downarrow$} & \textbf{MPJVE $\downarrow$} & 
\textbf{MPJPE $\downarrow$} & \textbf{MPJVE $\downarrow$} & 
\textbf{MPJPE $\downarrow$} & \textbf{MPJVE $\downarrow$} \\
\midrule
Ours w/o image & 1.34 & 0.68 & 1.56 & 0.44 & 2.85 & 0.60 \\
Ours w/o mesh  & 1.88 & 0.63 & 2.25 & 0.39 & 3.16 & 0.44 \\
Ours w/o GMHA  & 1.08 & 0.55 & 1.49 & 0.38 & 1.82 & 0.37 \\
\textbf{Ours}  & \textbf{1.06} & \textbf{0.44} 
                & \textbf{1.28} & \textbf{0.37} 
                & \textbf{1.76} & \textbf{0.36} \\
\bottomrule
\end{tabular}
}
\caption{Ablation study on input modalities and modules. Lower is better ($\downarrow$). Results are reported on the Truebones Zoo-test set across three generalization levels: seen, rare, and unseen species.}
\label{tab:ablation-mpjpe-mpjve}
\end{table}

\begin{table}[t]
\centering
\normalsize
\setlength{\tabcolsep}{3pt}
\resizebox{\columnwidth}{!}{
\begin{tabular}{l c|cc|cc|cc}
\toprule
\multirow{2}{*}{\textbf{Method}} &
\multirow{2}{*}{\textbf{Enc/Dec}} &
\multicolumn{2}{c|}{\textbf{Seen}} &
\multicolumn{2}{c|}{\textbf{Rare}} &
\multicolumn{2}{c}{\textbf{Unseen}} \\
\cmidrule(lr){3-4}
\cmidrule(lr){5-6}
\cmidrule(lr){7-8}
& &
\textbf{MPJPE $\downarrow$} & \textbf{MPJVE $\downarrow$} &
\textbf{MPJPE $\downarrow$} & \textbf{MPJVE $\downarrow$} &
\textbf{MPJPE $\downarrow$} & \textbf{MPJVE $\downarrow$} \\
\midrule
Variant 1 & 1 / 12 & 1.07 & 0.44 & 1.90 & 0.39 & 2.18 & 0.38 \\
Variant 2 & 2 / 12 & 1.11 & 0.51 & 1.37 & 0.38 & 2.00 & 0.37 \\
Variant 3 & 4 / 16 & \textbf{1.05} & \textbf{0.42} & 1.46 & 0.39 & 1.85 & 0.38 \\
\textbf{Ours} & \textbf{4 / 12} &
{1.06} & {0.44} &
\textbf{1.28} & \textbf{0.37} &
\textbf{1.76} & \textbf{0.36} \\
\bottomrule
\end{tabular}
}
\caption{
Ablation results with encoder/decoder layer configurations. 
Metrics are reported on the Truebones Zoo-test set under seen, rare, and unseen generalization.
}
\label{tab:ablation-architecture}
\end{table}

\subsection{Ablation Study}
We first analyze the impact of input modalities and key modules. Specifically, we consider variants that remove the reference image-set encoder and its cross-attention modules (w/o image), exclude mesh features from both reference and video streams (w/o mesh), and disable the graph multi-head attention over the skeleton (w/o GMHA), in comparison to the full model.
As shown in Table \ref{tab:ablation-mpjpe-mpjve}, removing any modality or module leads to clear performance drops, especially in the rare and unseen splits. The mesh and graph-attention branches are crucial for robust transfer to new species, highlighting the importance of explicit topology and geometry modeling.

Table~\ref{tab:ablation-architecture} further examines the impact of encoder and decoder layer configurations.
Increasing encoder and decoder depth generally improves performance, especially on rare and unseen species, indicating the importance of sufficient model capacity for handling diverse motion patterns.
Although Variant 3 achieves slightly better performance on the seen split, it introduces higher model complexity with limited gains on rare and unseen cases.
We therefore adopt a balanced configuration (4 encoder layers, 12 decoder layers), which achieves consistently strong performance across all splits.

\subsection{Qualitative Results on Truebones Zoo}

\begin{figure}[t]
    \centering
    \includegraphics[width=0.45\textwidth]{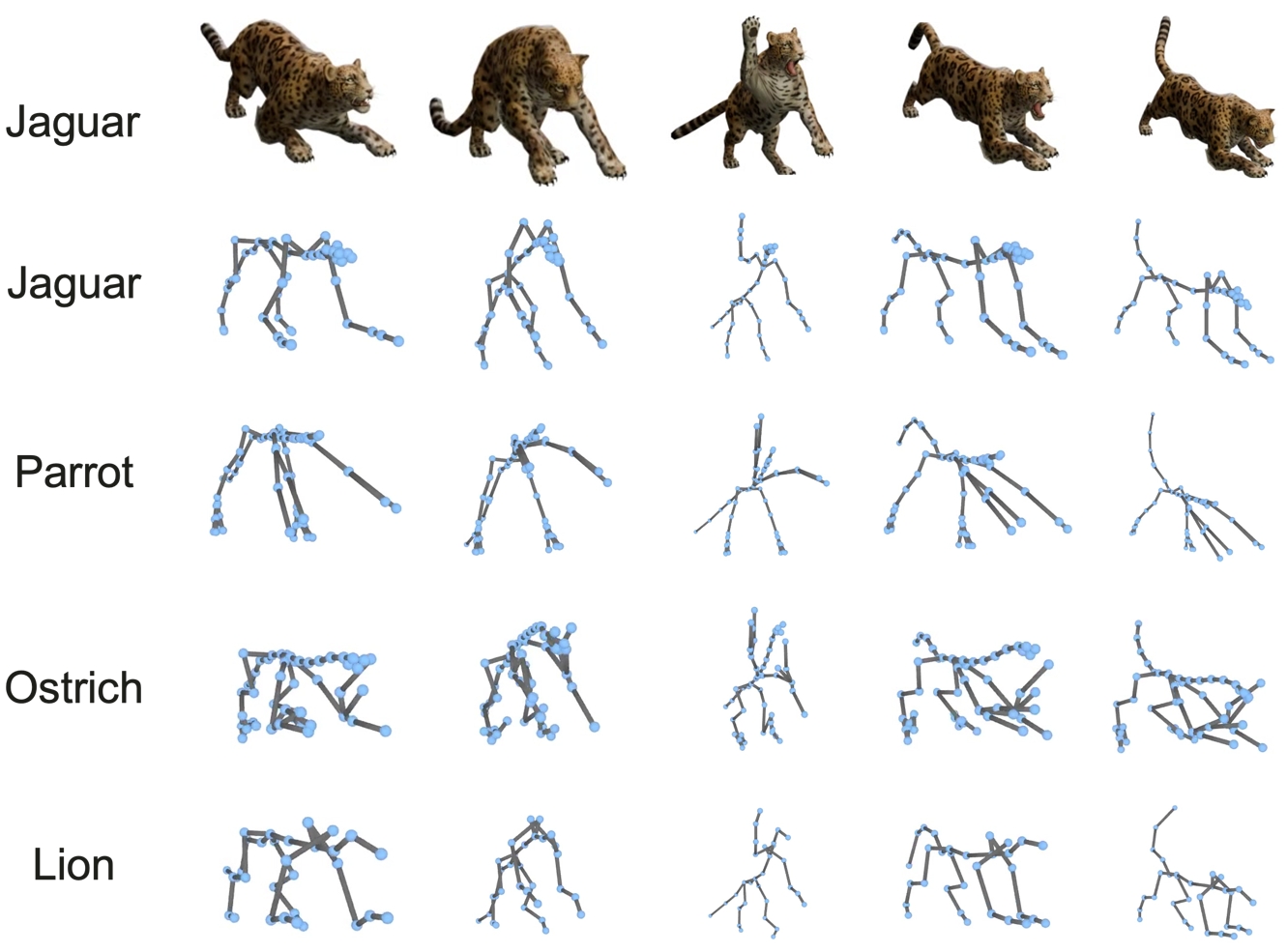}
    \caption{\textbf{Truebones Zoo mocap and retargeting results.}
    Each row visualizes one evaluation sequence.
    \textbf{Row 1}: Input video frames.
    \textbf{Row 2}: Same-species reference skeleton and predicted mocap results.
    \textbf{Rows 3--5}: Reference skeletons from three different species and retargeted motions by our method.
    Our method generalizes across species and produces stable, anatomically plausible 3D motion.}
    \label{fig:result-zoo}
\end{figure}

Figure~\ref{fig:result-zoo} presents representative Truebones Zoo-test results. 
\textbf{Row 1} shows input video Jugar. 
\textbf{Row 2} displays the same-species reference and predicted mocap outputs.
\textbf{Rows 3--5} show results when retargeting to skeletons of three different species.
Our approach generalizes robustly across species and maintains temporally consistent, anatomically plausible 3D motion even with significant appearance and shape variation.

\subsection{Qualitative Results on Objaverse}

In addition to animal skeletons, our framework also supports human-like rigs, enabling both human motion capture and cross-domain retargeting between humans and animals. Our model can transfer motion from humans to animals and vice versa, demonstrating strong versatility across different skeleton types. Representative qualitative results: including humanoid mocap, human to animal, and animal to human retargeting, are provided on our project homepage.


\begin{figure}[t]
    \centering
    \includegraphics[width=0.45\textwidth]{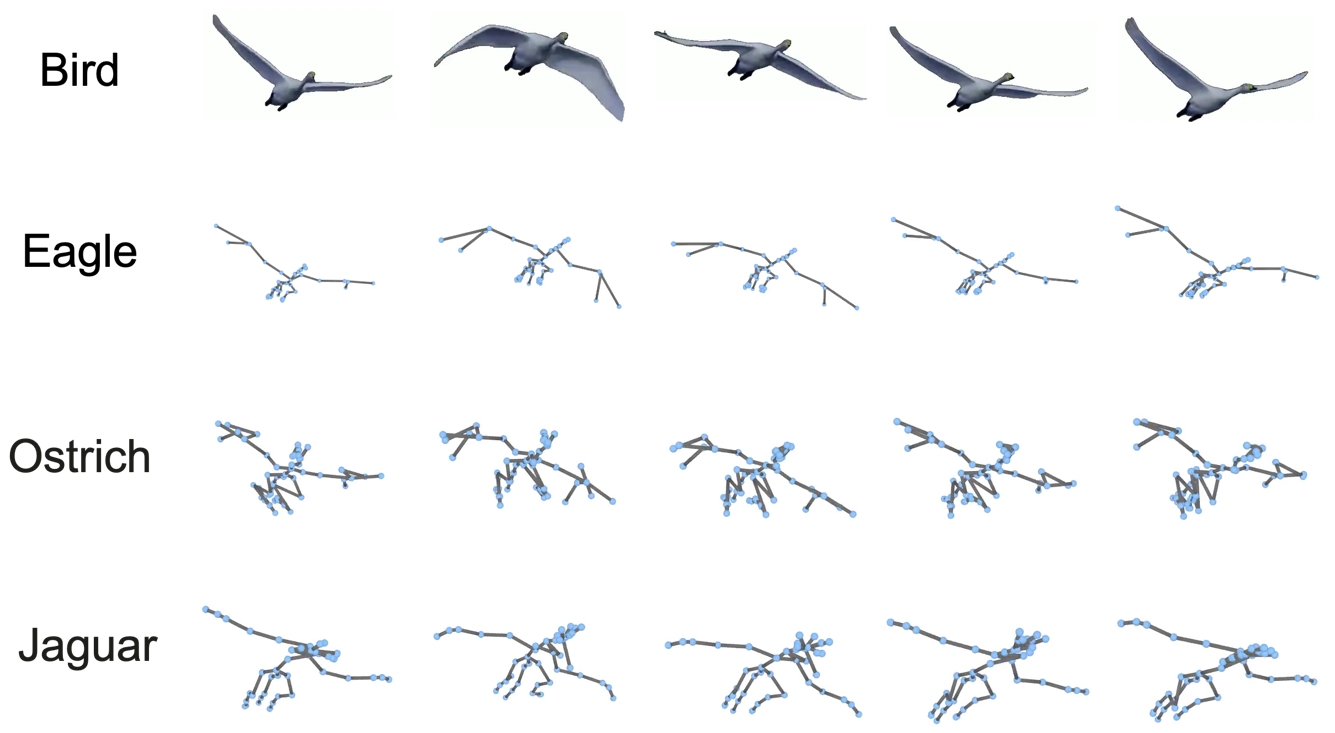}
    \caption{\textbf{Real-world (Wild) results.}
    Similar layout as Figure~\ref{fig:result-zoo}.
    \textbf{Row 1}: Input wild video frames.
    \textbf{Row 2}: Same-species reference skeleton and predicted mocap outputs.
    \textbf{Rows 3--4}: Cross-species reference skeletons and our retargeted motion predictions.
    Despite real-world challenges, our method maintains robustness and stability.}
    \label{fig:result-wild}
\end{figure}

\subsection{In-the-Wild Generalization}

To further assess robustness, we apply our trained model to a variety of in-the-wild animal videos collected from the Internet, including birds (chickens, eagles, seagulls), quadrupeds (tigers, leopards, elephants, cats, dogs), and other animals such as crabs, fish, and snakes. As shown in Figure~\ref{fig:result-wild}, our method successfully reconstructs plausible 3D skeletal motion for both mocap and retargeting (Jaguar pretend to fly), demonstrating strong generalization.

\subsection{Visualization of IK-Driven Asset Animation}

We further visualize the IK-driven animation results to assess the full pipeline from predicted skeletons to rigged assets. 
For each example, we show multi-view renderings of the predicted 3D skeletons together with the corresponding IK-driven asset animations.
As illustrated (see Figures~\ref{fig:ik-result}) in our examples—including an ostrich (biped), a goat (quadruped), and in-the-wild cases such as a crab and a dog. 
These results demonstrate that our approach generalizes well across different species and real-world scenarios, enabling reliable motion capture and animation beyond controlled settings.

\subsection{Arbitrary Cross-Species Retargeting}

A unique feature of our approach is \textbf{prompt-based retargeting} across arbitrary asset types: even for reference skeletons entirely unrelated to the subject in the input video. Although not explicitly trained for cross-species transfer, our model leverages structural, visual, and geometric cues to synthesize plausible retargeted motion. 

\begin{figure}[t]
    \centering
    \includegraphics[width=0.45\textwidth]{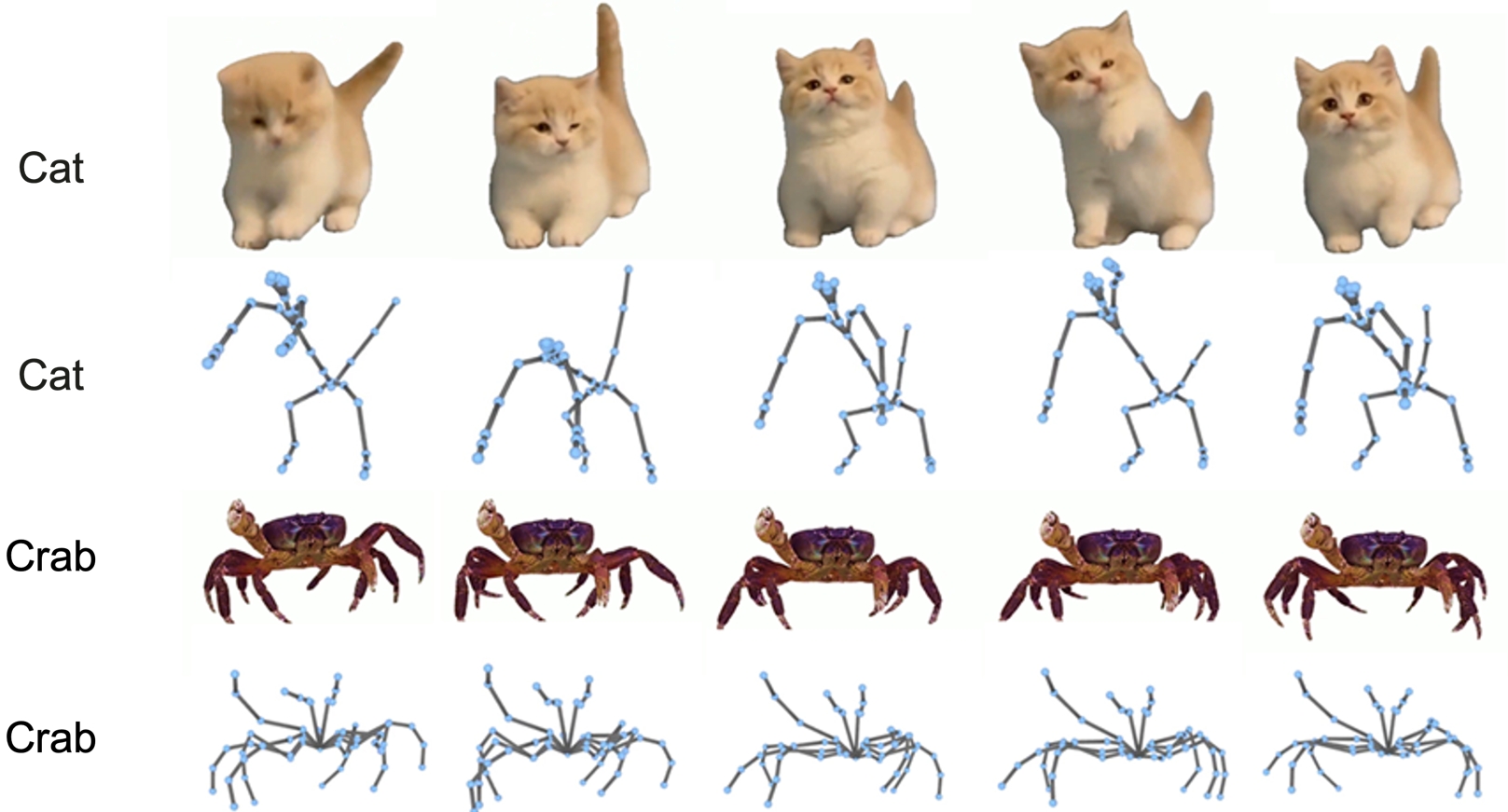}
    \caption{\textbf{More in-the-wild mocap results.} Our method generalizes to a diverse range of species and scenarios.}
    \label{fig:more-mocap}
\end{figure}

\begin{figure}[t]
    \centering
    \includegraphics[width=0.45\textwidth]{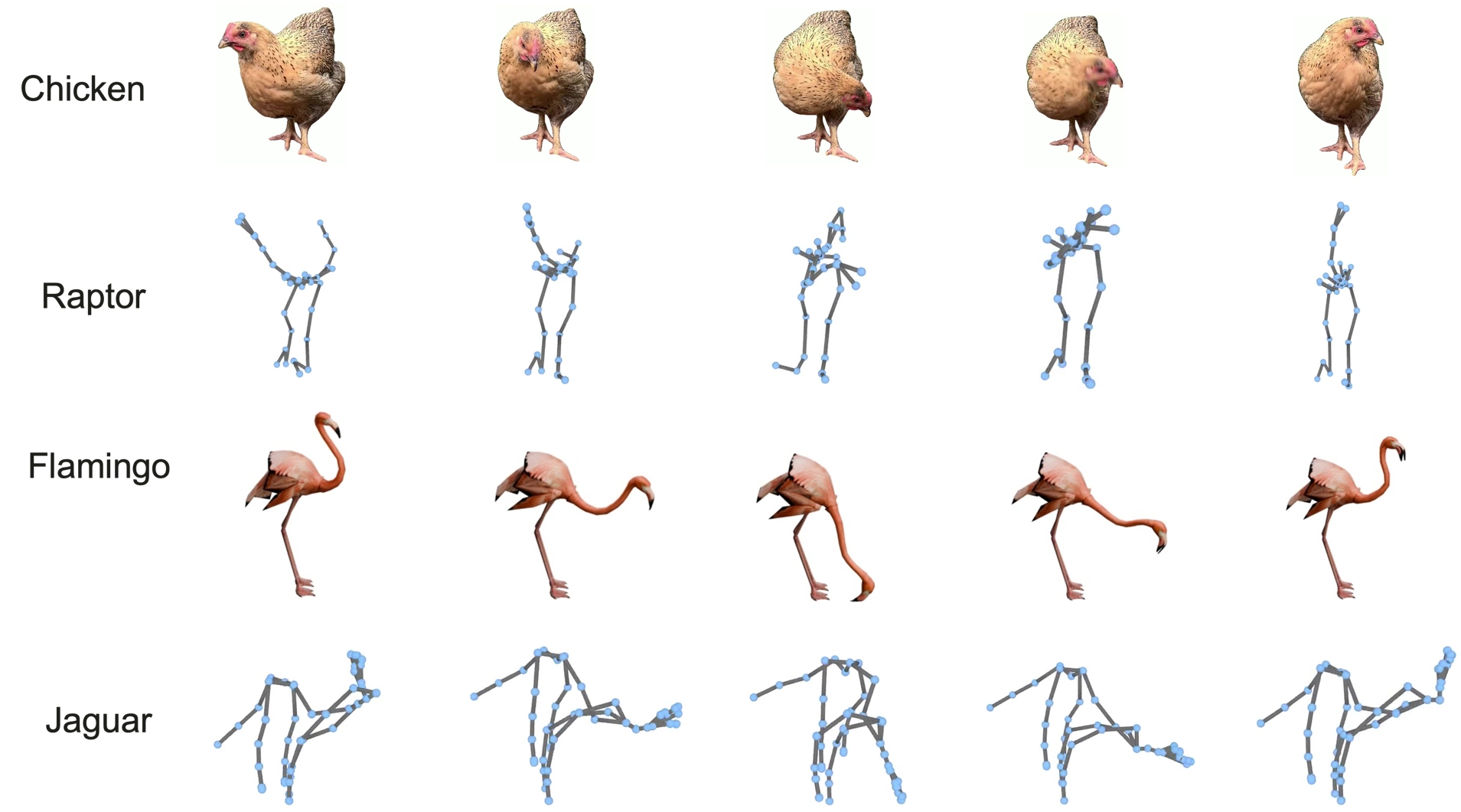}
    \caption{\textbf{Unconstrained cross-species retargeting.} Examples of using our model to retarget motion from one species to another, yielding diverse, creative, and physically plausible animations. 1st row: from chicken to Raptor, 2nd row: Flamingo to Jaguar.}
    \label{fig:more-retarget}
\end{figure}

We observe a wide range of creative results: bird videos drive quadrupeds to perform flapping-like actions or animate pterosaurs; fish swimming is transferred to crocodiles or snakes; dog running animates bipedal birds; crocodile tail-whipping is retargeted to leopards or parrots. 
Such unconstrained retargeting enables new workflows for animation (see Figures~\ref{fig:more-mocap} and~\ref{fig:more-retarget} for more results).

Given the lack of directly comparable baselines, we focus on extensive qualitative analysis and ablation, providing thorough visualization of our results and highlighting the practical versatility of our approach.


\section{Conclusion}
\label{sec:conclusion}

In this work, we reformulate the motion capture problem as Category-Agnostic Motion Capture (CAMoCap), a novel paradigm in which a monocular video and an arbitrary rigged asset function as input prompts to generate rotation-based animations tailored to the target character. We further propose \name, a reference-guided factorized architecture that initially estimates 3D joint trajectories and subsequently reconstructs asset-specific rotations through constraint-aware inverse kinematics, while mitigating cross-modal discrepancies between RGB and joint representations via an intermediate coarse 4D mesh. Using our proposed reorganized the \textbf{Truebones Zoo} benchmark, comprising 1,038 annotated clips with 60 test sequences and providing standardized skeleton-mesh-rendered video triples, \name consistently produces temporally stable, animation-ready outputs across diverse rigging systems, demonstrating notable in-domain precision, robust generalization to in-the-wild scenarios, and semantically meaningful cross-species motion retargeting capabilities.

\noindent\textbf{Limitations and future work.} Our performance depends on the quality of the pretrained image-to-3D reconstructor and assumes access to a rig with known joint structure; it also operates primarily in camera space without explicit physics or contact reasoning. Future directions include end-to-end, contact- and physics-aware IK, world-grounded trajectory recovery, reducing reliance on 4D reconstruction (e.g., video-only geometry priors), text-only or multimodal prompts beyond rendered images, and extensions to multi-character interaction.
{
    \small
    \bibliographystyle{ieeenat_fullname}
    \bibliography{main}
}

\clearpage
\setcounter{page}{1}
\maketitlesupplementary

\section{More Visualization Results}

In this section, we summarize additional qualitative results from our
\textbf{supplementary webpage}. These visualizations highlight the effectiveness
of our approach across controlled multi-species datasets, in-the-wild
videos, and cross-species retargeting scenarios, showing that our model
produces high-fidelity and temporally smooth motion under a broad range
of conditions.
\paragraph{Comparison with GenZoo.}
We compare our results with GenZoo, a single-image animal pose and shape
estimator trained on synthetic quadruped data. Without temporal modeling,
GenZoo exhibits frame-wise inconsistencies and pose fluctuations when
applied to video sequences, even for quadruped inputs. In contrast, our
method models motion dynamics explicitly, yielding smoother and more
coherent 4D reconstructions that better follow ground-truth trajectories.

\paragraph{Mocap Results.}
The supplementary webpage provides additional mocap visualizations.
From Truebones Zoo, we show examples spanning multiple animal species
with diverse skeletal structures; from Objaverse, we include bipedal
characters to demonstrate adaptability across different asset types. We
also present in-the-wild cases such as flying birds and crocodiles to
illustrate performance on real video inputs.

\paragraph{Arbitrary Motion Retargeting.}
We further include motion retargeting examples: Zoo2Zoo transfer across
different animal species, Human2Zoo transfer applying human motions to
animal skeletons, and Zoo2Human transfer mapping animal motions to a
human skeleton. For In-the-Wild2Human results, motions from videos of
animals such as eagles and leopards are retargeted to a human skeleton.
These examples show that our model handles large variations in
morphology, topology, and motion dynamics.

\paragraph{IK Visualization.}
We also provide IK fitting visualizations, showing recovered joint rotations and the improved temporal stability and orientation consistency achieved through geometric initialization, temporal warm-starting, and twist-regularized refinement. 
We additionally report an average geodesic rotation error of approximately 17$^\circ$, indicating reasonable rotation accuracy after IK.

\section{More Experiment Results}

\begin{table}[h]
\centering
\normalsize
\begin{tabular}{lccc}
\toprule
\textbf{Model} & \textbf{Quad} & \textbf{Non-Quad} & \textbf{All} \\
\midrule
\textbf{genzoo} & 0.4466 & 0.4740 & 0.4580 \\
\textbf{ours}   & 0.2354 & 0.2821 & 0.2549 \\
\bottomrule
\end{tabular}
\caption{
\normalsize Chamfer Distance (CD) results on the Truebones Zoo dataset.
}
\label{tab:metric-analysis}
\end{table}

\begin{figure}[t]
    \centering
    \includegraphics[width=0.45\textwidth]{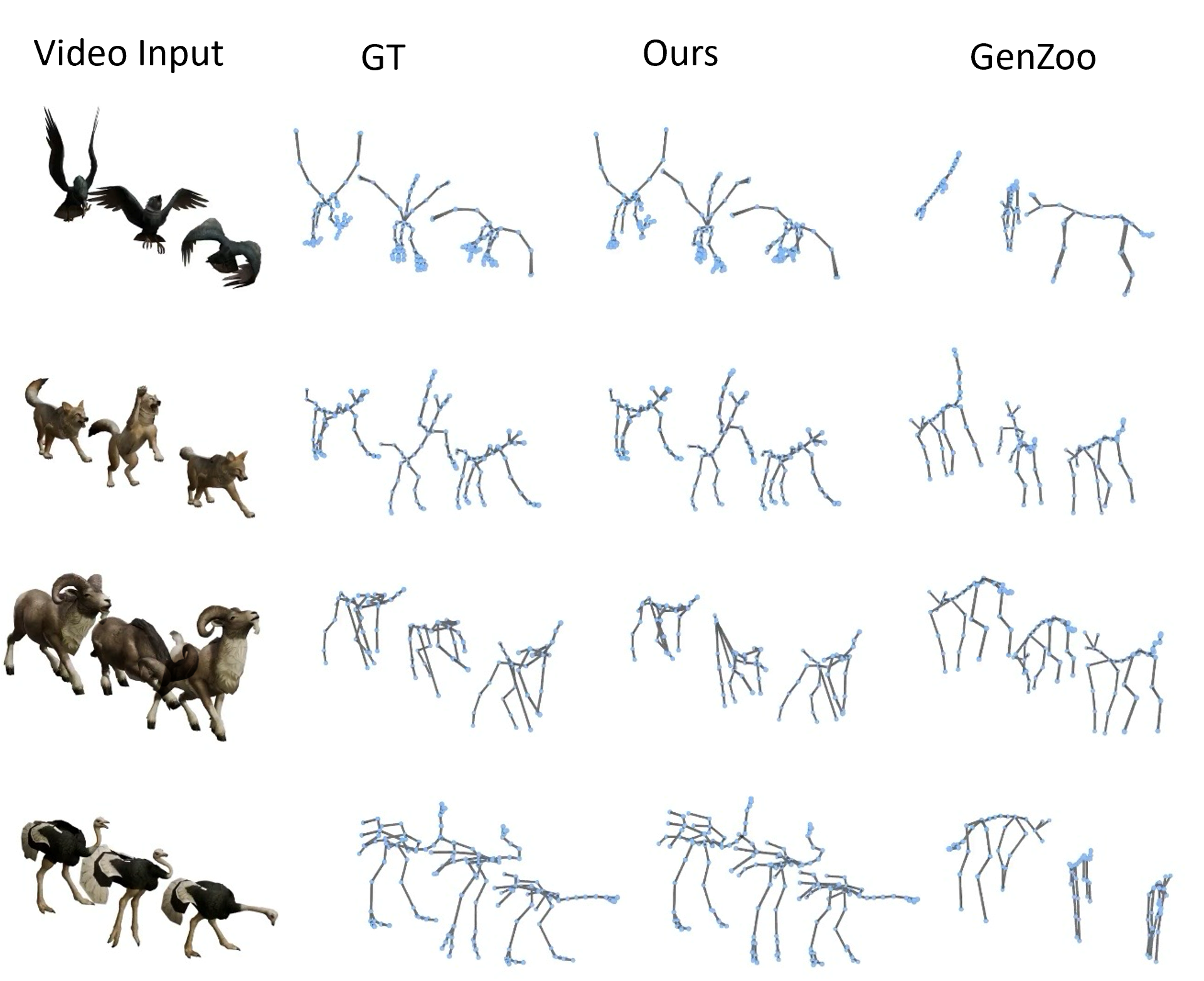}
    \caption{\textbf{Qualitative comparison with GenZoo on the Truebones Zoo dataset.}
    Our method produces smoother trajectories and maintains stable, anatomically plausible motions across a wide variety of skeleton types, including non-quadrupeds. In contrast, GenZoo is limited to quadruped structures and often fails to generalize to more diverse or complex skeletal configurations. Visualizations highlight our approach's superior accuracy, robustness, and generalization ability.}
    \label{fig:result-compare}
\end{figure}

To our knowledge, GenZoo~\cite{genzoo2025ICCV} is among the few works that attempt category-agnostic animal motion capture.
However, it mainly supports quadruped species and struggles to generalize to more diverse skeletons.
Since GenZoo does not produce joint-aligned skeletons compatible with MPJPE/MPJVE evaluation, 
we adopt \textbf{Chamfer Distance (CD-Skeleton)} as a structural metric for comparison.
For a comprehensive comparison, we evaluate both methods on the Truebones Zoo dataset, using the CD-Skeleton metric to measure the structural accuracy of the predicted skeletal motion.

As shown in Table~\ref{tab:metric-analysis}, our approach achieves significantly lower CD-Skeleton errors than GenZoo across all categories. On the overall test set, our method reduces the average error from 0.4580 to 0.2549, indicating a substantial improvement in capturing and reconstructing diverse skeletal motions, especially for non-quadruped species where existing methods perform poorly.

Figure~\ref{fig:result-compare} presents representative qualitative results on the Truebones Zoo dataset. Compared to GenZoo, our predictions exhibit smoother motion trajectories, higher anatomical fidelity, and robust stability across both quadruped and non-quadruped skeletons—including bipeds, birds, reptiles, and even non-biological assets. GenZoo, while currently the most widely applicable animal motion capture method, is fundamentally constrained by its reliance on quadruped skeleton templates and struggles to generalize to broader categories.

For further qualitative comparison, we provide side-by-side visualizations of our results and GenZoo's on our project homepage, showcasing the advantages of our approach in both accuracy and generalization.

\section{Implementation Details}

\subsection*{A. Dataset and Training Details}

\paragraph{Dataset Processing.}
Our 60-sequence benchmark composition: 27 mammals, 9 birds, 12 dinosaurs+dragons, 7 reptiles (incl. snakes), 2 aquatic, 3 arthropods. All meshes and joints are first scaled by the bounding box of their rest pose, normalizing each mesh into a unit-volume space. For sequence data, we remove the global translation of every frame, compute a sequence-level super bounding box, and uniformly scale the entire sequence into the range $[-1, 1]^3$. For in-the-wild video inputs, we assume a fixed camera position throughout the sequence.

\paragraph{Training details.}
The network consists of 12 layers for decoder, and a prompt encoder composed of 4 layers. All experiments are conducted on 8 GPUs, each equipped with 64\,GB of memory. The model is trained for 60 epochs using the Adam optimizer, requiring approximately 36 hours in total. We use a learning rate of $1\times10^{-4}$ and a batch size of 1 per GPU.
Training is performed with paired supervision for motion capture (not retargeting). 
For each sample, we select a reference asset from the same species (one frame providing image, unordered mesh, and skeleton as prompt) and predict 3D joint positions for a 24-frame input video. 
The loss is defined on joint positions, followed by a lightweight inverse kinematics (IK) fitting step to recover joint rotations for deployment.
We employ a sliding-window mechanism to support inference on arbitrarily long videos. 
During attention computation, masked joints are excluded, and both joint identity embeddings and skeletal topology are incorporated as conditioning signals.
We do not explicitly train on retargeting pairs. Nevertheless, the learned reference-conditioned motion representation enables cross-species retargeting behaviors at inference time.
During training, we use ground-truth mesh sequences for efficiency. At inference, we replace them with meshes predicted by a video-to-mesh module (e.g., SWIFT4D), which provide sufficiently stable conditioning in practice, and we empirically observe negligible degradation compared to GT-mesh conditioning.

\subsection*{B. Inverse Kinematics Fitting}

Given a predicted sequence of 3D joint locations 
$\{\mathbf{X}_{t,i}\}$ and a kinematic tree with rest-pose
offsets $\mathbf{o}_i$ and parent indices $p(i)$, our goal is to
recover temporally stable joint rotations 
$\mathbf{R}_{t,i}\!\in\!SO(3)$ such that the forward kinematics (FK)
matches the observed joints:
\[
\mathbf{P}_{t,i}=
\begin{cases}
\mathbf{0}, & p(i)=-1, \\
\mathbf{P}_{t,p(i)} + \mathbf{R}_{t,p(i)}\,\mathbf{o}_i,
& \text{otherwise}.
\end{cases}
\]

Because FK is not injective, position-only constraints do not fully
determine local orientation, especially twist around the bone axis.
We therefore combine geometric initialization, temporal warm-starting,
and differentiable refinement with twist suppression.

\vspace{3pt}
\paragraph{Geometric Initialization.}
For each frame, we compute a closed-form IK estimate 
$\mathbf{R}^{\text{geo}}_{t,i}$.
For single-child joints, we align rest-pose and observed bone vectors
via axis--angle rotation.
For multi-child joints, we solve the orthogonal Procrustes problem:
\[
\mathbf{R}^{\text{geo}}_{t,i}
=
\arg\min_{\mathbf{R}\in SO(3)}
\sum_{k}
\big\|
\mathbf{R}\,\mathbf{v}^{\text{rest}}_{i,k}
-
\mathbf{v}^{\text{obs}}_{t,i,k}
\big\|^2,
\]
where $\mathbf{v}^{\text{rest}}$ are rest-space bone directions and
$\mathbf{v}^{\text{obs}}$ are normalized directions from predicted joints.
This provides consistent orientations at branching structures 
(e.g., pelvis, shoulders).

\vspace{3pt}
\paragraph{Temporal Warm-Starting.}
To avoid frame-to-frame drift, optimization for frame $t$
is initialized using the solution from the previous frame:
\[
\boldsymbol{\theta}^{(0)}_t = \boldsymbol{\theta}^{\ast}_{t-1}.
\]

\vspace{3pt}
\paragraph{Differentiable Refinement.}
Local rotations are parameterized as axis--angle vectors
$\boldsymbol{\theta}_{t,i}\!\in\!\mathbb{R}^{3}$ and refined via the loss:
\[
\mathcal{L}_t
=
\mathcal{L}_{\text{pos}}
+
\lambda_{\text{prior}}\mathcal{L}_{\text{prior}}
+
\lambda_{\text{twist}}\mathcal{L}_{\text{twist}}.
\]

The FK position loss is:
\[
\mathcal{L}_{\text{pos}}
=
\frac{1}{N}
\sum_{i}
\big\|
\mathbf{P}_{t,i}(\boldsymbol{\theta}_t) - \mathbf{X}_{t,i}
\big\|^2.
\]

A geometric prior encourages solutions close to the
closed-form initialization:
\[
\mathcal{L}_{\text{prior}}
=
\frac{1}{N}
\sum_{i}
\|\boldsymbol{\theta}_{t,i} 
- 
\boldsymbol{\theta}^{\text{geo}}_{t,i}\|^2.
\]

\vspace{2pt}
\paragraph{Twist Suppression.}
Since bone-axis twist is under-constrained, we penalize rotation
components parallel to the bone direction 
$\mathbf{u}_i = \mathbf{o}_i /\|\mathbf{o}_i\|$.
Let
$\boldsymbol{\theta}_{t,i}=\alpha_{t,i}\hat{\mathbf{a}}_{t,i}$.
The twist magnitude is:
\[
\alpha^{\text{twist}}_{t,i}
=
\alpha_{t,i}\,
(\hat{\mathbf{a}}_{t,i}\!\cdot\!\mathbf{u}_i).
\]
We minimize:
\[
\mathcal{L}_{\text{twist}}
=
\frac{1}{N}
\sum_i
\big(\alpha^{\text{twist}}_{t,i}\big)^2.
\]

This term suppresses candy-wrapper artifacts while preserving natural
motion around long chains such as tails.

\vspace{3pt}
\paragraph{Summary.}
The combination of geometric IK, temporal warm-starting, and
twist-regularized refinement yields stable and anatomically consistent
joint rotations, significantly improving reconstruction quality.
Further implementation details are provided in the code release.


\section{Evaluation Metrics}

This section describes the computation of the proposed metric(CD-Skeleton) that evaluates the alignment between two articulated skeletons. Each skeleton is represented by a set of 3D joint positions and a kinematic hierarchy defined by a parent array.

\subsection*{Notation}

Let Skeleton A and Skeleton B be defined as:
\begin{itemize}
    \item Joint positions:
    \[
    \begin{aligned}
    \mathbf{X}^{A} 
    &= \{\mathbf{x}^{A}_{i} \in \mathbb{R}^3 \mid i = 1,\dots,N\}, 
    \\
    \mathbf{X}^{B} 
    &= \{\mathbf{x}^{B}_{i} \in \mathbb{R}^3 \mid i = 1,\dots,N\}.
    \end{aligned}
    \]

    where $N$ is the number of joints.
    \item Kinematic hierarchy, defined by a parent array:
    \[
        \mathbf{p}^{A}, \mathbf{p}^{B} \in \{-1,1,\dots,N\}^N,
    \]
    where $p^{A}_i = -1$ (or $p^B_i = -1$) indicates a root joint.
\end{itemize}

Although the parent arrays may differ, the metric assumes a known correspondence of joint indices between the two skeletons.

\subsection*{Distance From Joint to the Other Skeleton}

For each joint of Skeleton A, we compute its distance to the closest point on the bone segments of Skeleton B. Skeleton B consists of line segments defined by its kinematic tree:

\[
    \mathcal{S}^B = \{ (\mathbf{x}^B_{i},\, \mathbf{x}^B_{p^B_i}) \mid p^B_i \neq -1 \}.
\]

For a joint $\mathbf{x}^A_{i}$, its distance to Skeleton B is defined as:
\[
    d(\mathbf{x}^A_i, \mathcal{S}^B)
    =
    \min_{( \mathbf{b}_1,\mathbf{b}_2 ) \in \mathcal{S}^B }
    \left\|
        \mathbf{x}^A_i - \Pi_{\mathbf{b}_1,\mathbf{b}_2}(\mathbf{x}^A_i)
    \right\|,
\]
where $\Pi_{\mathbf{b}_1,\mathbf{b}_2}(\mathbf{v})$ denotes the orthogonal projection of point $\mathbf{v}$ onto the line segment connecting $\mathbf{b}_1$ and $\mathbf{b}_2$. This projection is computed as:
\[
    \Pi_{\mathbf{b}_1,\mathbf{b}_2}(\mathbf{v})
    = 
    \mathbf{b}_1
    +
    \text{clip}\!\left(
        \frac{(\mathbf{v} - \mathbf{b}_1)\cdot(\mathbf{b}_2 - \mathbf{b}_1)}
             {\|\mathbf{b}_2 - \mathbf{b}_1\|^2},
        0,\, 1
    \right)
    (\mathbf{b}_2 - \mathbf{b}_1),
\]
where $\text{clip}(t,0,1) = \max(0, \min(t,1))$ ensures the projected point lies on the segment.

Similarly, we can compute the distance from joints of Skeleton B to Skeleton A.

\subsection*{Skeleton-to-Skeleton Distance}

The asymmetric distance from Skeleton A to Skeleton B is:
\[
    D(A \rightarrow B)
    =
    \frac{1}{N}
    \sum_{i=1}^{N}
        d(\mathbf{x}^A_i, \mathcal{S}^B).
\]

The symmetric distance is defined as:
\[
    D_{\mathrm{sym}}(A,B)
    =
    \frac{1}{2}\left(
        D(A \rightarrow B) + D(B \rightarrow A)
    \right).
\]

\subsection*{Interpretation}

This metric evaluates how closely each joint of one skeleton lies to the structure of the other skeleton, capturing differences in global pose, limb orientation, and proportions. The symmetric version provides a balanced measure when neither skeleton should be considered the reference.

\end{document}